%% file: iclr2020_conference.tex
\documentclass{article} 
\usepackage{iclr2020_conference,times}

\input{math_commands.tex}

\usepackage{hyperref}
\usepackage{url}
\usepackage[pdftex]{graphicx}
\usepackage{enumitem}
\usepackage{cleveref}
\usepackage{pifont}
\usepackage{subcaption}
\usepackage{chngpage}
\usepackage{xcolor}

\definecolor{green}{rgb}{0.122, 0.616, 0.353}

\hypersetup{
    colorlinks = true,
    urlcolor = blue,
    citecolor=.,
    linkcolor=.,
}

\makeatletter
\def\thickhline{%
  \noalign{\ifnum0=`}\fi\hrule \@height \thickarrayrulewidth \futurelet
   \reserved@a\@xthickhline}
\def\@xthickhline{\ifx\reserved@a\thickhline
               \vskip\doublerulesep
               \vskip-\thickarrayrulewidth
             \fi
      \ifnum0=`{\fi}}
\makeatother

\newlength{\thickarrayrulewidth}
\title{Learning a Spatio-Temporal Embedding\\for Video Instance Segmentation}

\author{Anthony Hu, Alex Kendall \& Roberto Cipolla\\
University of Cambridge\\
\texttt{\{ah2029,agk34,rc10001\}@cam.ac.uk}
}

\iclrfinalcopy 
\begin{document}

\maketitle

\begin{abstract}
We present a novel embedding approach for video instance segmentation. Our method learns a spatio-temporal embedding integrating cues from appearance, motion, and geometry; a 3D causal convolutional network models motion, and a monocular self-supervised depth loss models geometry. In this embedding space, video-pixels of the same instance are clustered together while being separated from other instances, to naturally track instances over time without any complex post-processing. Our network runs in real-time as our architecture is entirely causal -- we do not incorporate information from future frames, contrary to previous methods. We show that our model can accurately track and segment instances, even with occlusions and missed detections, advancing the state-of-the-art on the KITTI Multi-Object and Tracking Dataset.
\end{abstract}

\section{Introduction}
Explicitly predicting the motion of actors in a dynamic scene is a critical component of intelligent systems.
Humans can seamlessly track moving objects in their environment by using cues such as appearance, relative distance, and most of all, temporal consistency: the world is rarely experienced in a static way: motion (or its absence) provides essential information to understand a scene. Similarly, incorporating past context through a temporal model is essential to segment and track objects consistently over time and through occlusions. 

\begin{figure}[b]
\begin{center}
\includegraphics[width=\linewidth]{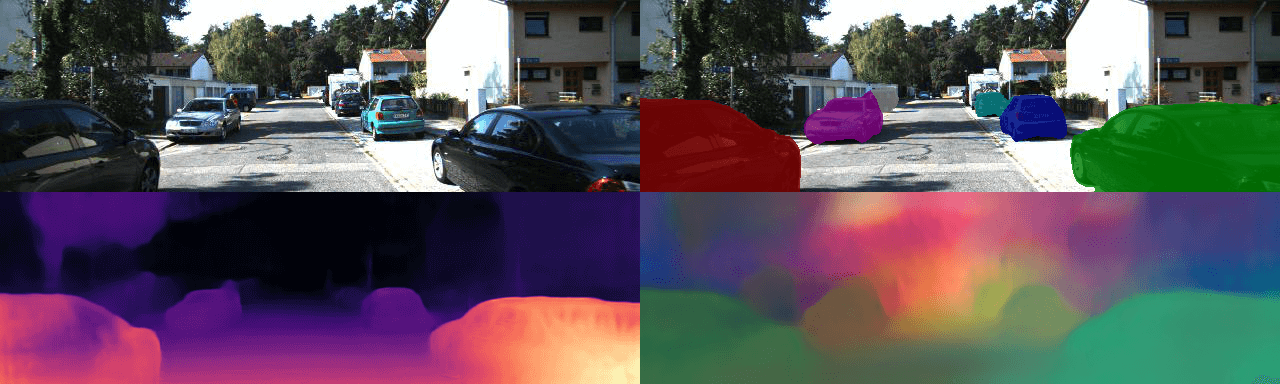}
\end{center}
\vspace{-3mm}
\caption{An illustration of our video instance segmentation model. Clockwise from top left: input image, predicted multi-object instance segmentation, visualisation of the high-dimensional embedding and predicted monocular depth.}
\label{fig:intro-figure}
\end{figure}

From a computer vision perspective, understanding object motion involves segmenting instances, estimating depth, and tracking instances over time. Instance segmentation has gained traction with challenging datasets such as COCO~\citep{coco14}, Cityscapes~\citep{cityscapes16} and Mapillary Vistas~\citep{mapillary17}. Such datasets, which only contain single-frame annotations, do not allow the training of video models with temporally consistent instance segmentation, nor does it allow self-supervised monocular depth estimation, that necessitates consecutive frames. Yet, navigating in the real-world requires temporally consistent segmentation and 3D gometry understanding of the other agents. More recently, a new dataset containing video instance segmentation annotations was released: the KITTI Multi-Object and Tracking Dataset \citep{voigtlaender19}. This dataset contains pixel-level instance segmentation on more than 8,000 video frames which effectively enables the training of video instance segmentation models. 

In this work, we propose a new spatio-temporal embedding loss that learns to map video-pixels to a high-dimensional space\footnote{See a \href{https://youtu.be/dc-3meFF6z0}{video demo} of our model.}. This space encourages video-pixels of the same instance to be close together and distinct from other instances. We show that this spatio-temporal embedding loss, jointly with a deep temporal convolutional neural network and self-supervised depth loss, produces consistent instance segmentations over time. The temporal model is a causal 3D convolutional network (only conditioned on past frames to predict the current embedding) and is capable of real-time operation. Finally, we show that predicting depth improves the quality of the embedding as the 3D geometry of an object constrains its future location given that objects move smoothly in space.

To summarise our novel contributions, we:

\begin{itemize}[noitemsep,topsep=0pt]
    \item introduce a new spatio-temporal embedding loss for video instance segmentation,
    \item show that having a temporal model improves embedding consistency over time,
    \item improve how the embedding disambiguates objects with a self-supervised monocular depth loss,
    \item handle occlusions, contrary to previous IoU based instance correspondence.
\end{itemize}
We demonstrate the efficacy of our method by advancing the state-of-the-art on the KITTI Multi-Object and Tracking Dataset~\citep{voigtlaender19}. An example of our model's output is given by \Cref{fig:intro-figure}.

\section{Related Work}
Two main approaches exist for single-image instance segmentation: region-proposal based \citep{he17,hu17,chen17b,liu18} and embedding based \citep{brabandere17,fathi17,kong18,kendall2017multi}. The former method relies on a region of interest proposal network that first predicts bounding boxes then estimates the mask of the object inside that bounding box. With such a strategy, a given pixel could belong to the overlap of many bounding boxes, and it is largely unclear how correspondence between pixels can be learned. We instead favour the embedding based method and extend it to space and time.

Capturing the inter-relations of objects using multi-modal cues (appearance, motion, interaction) is difficult, as showcased by the Multi-Object Tracking (MOT) challenge \citep{xiang15}. \citet{sadeghian17} and \citet{son17} learned a representation of objects that follows the "tracking-by-detection" paradigm where the goal is to connect detections across video frames by finding the optimal assignment of a graph-based tracking formulation (i.e. each detection is a node, and an edge is the similarity score between two detections).

Collecting large-scale tracking datasets is necessary to train deep networks, but that process is expensive and time-consuming. \citet{vondrick18} introduced video colourisation as a self-supervised method to learn visual tracking. They constrained the colourisation problem of a grayscale image by learning to copy colours from a reference frame, with the pointing mechanism of the model acting as a tracker once it is fully trained. The colourisation model is more robust than optical flow based models, especially in complex natural scenes with fast motion, occlusion and dynamic backgrounds.

\citet{voigtlaender19} extended the task of multi-object tracking to multi-object tracking and segmentation (MOTS), by considering instance segmentations as opposed to 2D bounding boxes. Motivated by the saturation of the bounding box level tracking evaluations \citep{ponttuset17}, they introduced the KITTI MOTS dataset, which contains pixel-level instance segmentation on more than 8,000 video frames. They also trained a model which extends Mask R-CNN \citep{he17} by incorporating 3D convolutions to integrate temporal information, and the addition of an association head that produces an association vector for each detection, inspired from person re-identification \citep{beyer17}. The temporal component of their model, however, is fairly shallow (one or two layers), and is not causal, as future frames are used to segment past frames. More recently, \citet{yang19} collected a large-scale dataset from short YouTube videos (3-6 seconds) with video instance segmentation labels, and \citet{hu19} introduced a densely annotated synthetic dataset with complex occlusions to learn how to estimate the spatial extent of objects beyond what is visible.

\section{Embedding-Based Video Instance Segmentation Loss}

Contrary to methods relying on region proposal \citep{he17,chen17b}, embedding-based instance segmentation methods map the pixels of a given instance to a structured high dimensional space, overcoming several limitations of region-proposal methods: (i) each pixel belongs to one unique instance (no bounding box overlap); (ii) the number of detected objects can be arbitrarily large (not fixed by the number of proposals).

We propose a spatio-temporal embedding loss with three competing forces, similarly to \citet{brabandere17}. The attraction force (\Cref{loss1}) encourages the video-pixels embedding of a given instance to be close to its embedding mean. The repulsion force (\Cref{loss2}) incites the embedding mean of a given instance to be far from all others instances. And finally, the regularisation force (\Cref{loss3}) prevents the embedding to diverge from the origin.

Let us denote by K the number of instances, and by $S_k$ the set of all video-pixels of instance $k$. For all $i \in S_k$, we denote by $y_i$ the embedding for pixel $i$ and by $\mu_k$ the mean embedding of instance $k$: $\mu_k = \frac{1}{|S_k|} \sum_{i \in S_k} y_i$. The embedding loss is given by:

\begin{align}
    \label{loss1}
	\mathcal{L}_{\text{a}} ={}& \frac{1}{K} \sum_{k=1}^K \frac{1}{|S_k|} \sum_{i \in S_k} \max(0, \Vert \mu_k - y_i \Vert_2 - \rho_a)^2\\
	\label{loss2}
	\mathcal{L}_{\text{r}} ={}& \frac{1}{K(K-1)} \sum_{k_1 \neq k_2} \max(0, 2\rho_r - \Vert \mu_{k_1} - \mu_{k_2} \Vert_2)^2\\
	\label{loss3}
	\mathcal{L}_{\text{reg}} ={}& \frac{1}{K} \sum_{k=1}^K \Vert \mu_k\Vert_2
\end{align}

$\rho_a$ defines the attraction radius, constraining the embedding to be within $\rho_a$ of its mean. $\rho_r$ is the repulsion radius, constraining the mean embedding of two different instances to be at least $2\rho_r$ apart. Therefore, if we set $\rho_r > 2\rho_a$, a pixel embedding of an instance $k$ will be closer to all the pixel embeddings $i \in S_k$ of instance $k$, than to the pixel embeddings of any other instance.

The spatio-temporal embedding loss is the weighted sum of the attraction, repulsion and regularisation forces:

\begin{equation}
	\mathcal{L}_{\text{instance}} = \lambda_a\mathcal{L}_{\text{a}} + \lambda_r\mathcal{L}_{\text{r}} + \lambda_{\text{reg}}\mathcal{L}_{\text{reg}}
\end{equation}

During inference, each pixel of the considered frame is assigned to an instance by randomly picking an unassigned pixel and aggregating close-by pixels with the mean shift algorithm \citep{comaniciu2002mean} until convergence. In the ideal case, with a test loss of zero, this will result in perfect instance segmentation.

\subsection{Self-Supervised Depth Estimation}
The relative distance of objects is a strong cue for segmenting instances in video. Knowing the 3D geometry of objects especially helps segmenting instances in a temporally consistent way, as the past position of an instance effectively constrains where it could be next.

Depth estimation with supervised methods requires a vast quantity of high quality annotated data, which is challenging to acquire in a range of environments. As we have access to a video instance segmentation dataset, we can use a self-supervised depth loss from monocular video, where the supervision comes from consecutive frames.

Following \citet{zhou17} and \citet{godard19}, we train a depth network with a separate pose estimation network with the hypothesis during training that scenes are mostly rigid, therefore assuming appearance change is mostly due to camera motion. Pixels that violate this assumption are masked from the view synthesis loss, as they otherwise create infinite holes during inference for objects that are typically seen in motion during training -- more details in \Cref{appendix-arch}. The training signal comes from novel view synthesis, i.e. the generation of a new image of the scene from a different camera pose. Let us denote by $(I_1, I_2, ..., I_T)$ a sequence of images, $I_t$ the target view and $I_s$ the source view. The view synthesis loss is given by:

\begin{equation}
\mathcal{L}_{vs} = \sum_{s \neq t} e(I_t, \hat{I}_{s\rightarrow t})
\end{equation}

with $\hat{I}_{s\rightarrow t}$ the synthesised view of $I_t$, from source image $I_s$ and using the predicted depth $\hat{D}_t$ and the estimated camera transformation $\hat{T}_{t\rightarrow s}$. The projection error $e$ is a weighted sum of an $\text{L}_1$ distance, a Structural Similarity Index (SSIM) and a smoothness regularisation term, as in \citet{zhao17}.
Let us denote by $p_t$ the coordinate of a pixel in the target image $I_t$ in homogeneous coordinates. Given the camera intrinsic matrix, $K$, and the mapping $\varphi$ from image plane to camera coordinate, the corresponding pixel in the source image is provided by: 

\begin{equation}
p_s \sim K\hat{T}_{t\rightarrow s}\varphi(K^{-1}p_t, \hat{D_t}(p_t))
\end{equation}

Since the projected coordinates $p_s$ take continuous values, we use the Spatial Transformer Network \citep{jaderberg15} sampling mechanism to bilinearly interpolate the four neighbouring pixels to populate the reconstructed image $\hat{I}_{s\rightarrow t}$.

Some pixels are visible in the target image, but are not in the source image, leading to a large projection error. As advocated by \citet{godard19}, instead of summing, taking the minimum projection error greatly reduces artifacts due to occlusion and results in sharper predictions. The resulting view synthesis loss is:

\begin{equation}
\mathcal{L}_{vs} = \min_{s \neq t} e(I_t, \hat{I}_{s\rightarrow t})
\end{equation}

The final video instance embedding loss is the weighted sum of the attraction, repulsion, regularisation and geometric view synthesis losses:

\begin{equation}
	\mathcal{L}_{\text{embedding}} = \lambda_a\mathcal{L}_{\text{a}} + \lambda_r\mathcal{L}_{\text{r}} + \lambda_{\text{reg}}\mathcal{L}_{\text{reg}} + \lambda_{vs}\mathcal{L}_{vs}
\end{equation}

\section{Model Architecture}
\begin{figure}[t]
\begin{center}
\includegraphics[width=\linewidth]{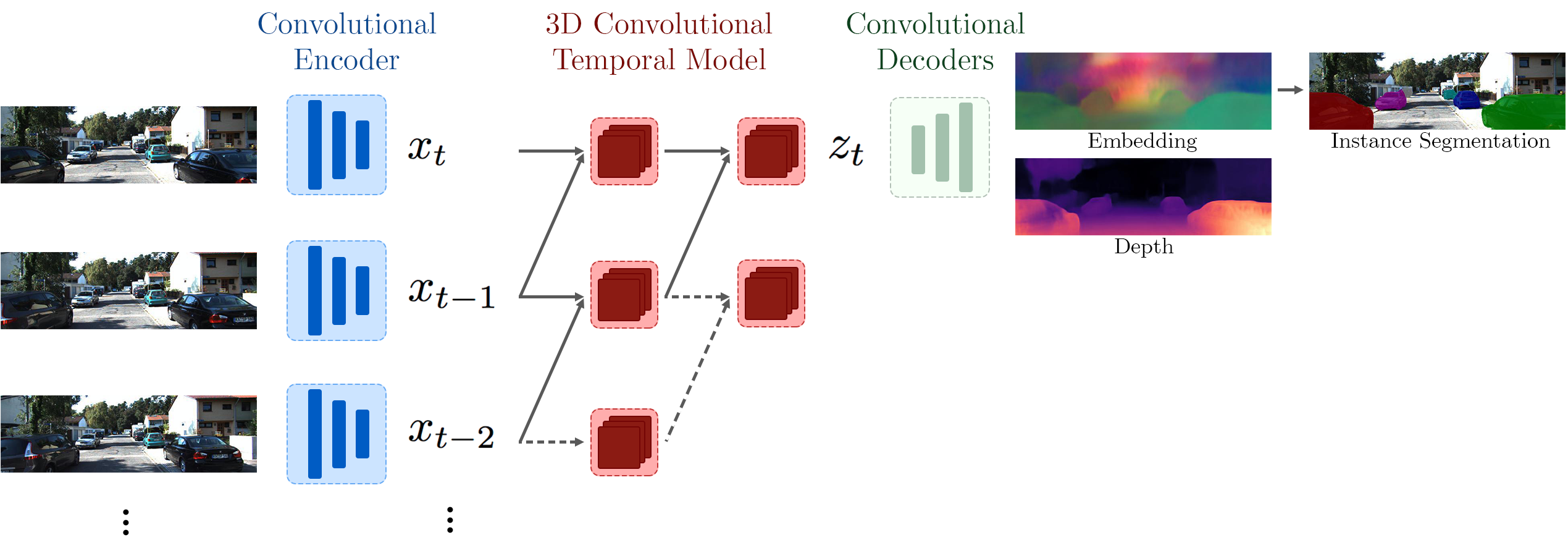}
\end{center}
\caption{Our spatio-temporal embedding network. The representation $z_t$ is trained to encode appearance, motion and geometry cues in order to predict instance embedding and monocular depth.}
\label{fig:model-architecture}
\end{figure}

Our network contains three components: the encoder, the temporal model and the decoders. Each input frame $I_t$ is first encoded as a compact feature $x_t$, then the temporal model learns a rich spatio-temporal representation $z_t$, and finally, the decoders output the instance embedding and depth prediction as illustrated by \Cref{fig:model-architecture}.

\paragraph{Encoder.}
We use a ResNet-18 \citep{he16} as our encoder, which allows the network to run in real-time on sequences of images.

\paragraph{Temporal Model.}
The model learns scene dynamics with a causal 3D convolutional network made of 3D residual convolutional blocks (convolving in both space and time, with residual connections). For a given time index, $t$, the network only convolves over images from indices $s \leq t$ to compute the temporal representation $z_t$. It therefore does not use future frames and is completely causal. The temporal model does not decimate the spatial dimension of the encoding, but slowly accumulates information over time from the previous encodings $x_s$ with $s \leq t$.
It is trained efficiently with convolutions as all input images are available during training, enabling parallel computations with GPUs. However, during inference, the model is inherently sequential, but can be made significantly faster by caching the convolutional features over time and eliminating redundant operations, as proposed by \citet{paine16} for WaveNet \citep{oord16}.

\paragraph{Decoders.}
From the temporal representation $z_t$, the decoders output the instance embedding $y_t \in \mathbb{R}^{p\times H \times W}$ and estimated depth $\hat{D}_t\in \mathbb{R}^{1\times H \times W}$, with $p$ the embedding dimension and $(H,W)$ the input image size.

\paragraph{Pose and Mask Model.}
The architecture of the Pose and Mask networks are given in \Cref{appendix-arch}.

\subsection{Inference}
For each new frame, we first mask the background with our mask network, then we cluster the foreground embeddings with mean shift to discover dense regions with each cluster corresponding to one instance. Tracking instances simply requires comparing the mean embedding of a newly segmented instance with previously segmented instances. A distance lower than $\rho_r$ indicates a match. 

The embeddings are accumulated over time, creating increasingly denser regions over time and resulting in a better clustering. To ensure that the pixel embeddings of a particular instance can smoothly vary over time, the embeddings have a life span corresponding to the time sequence length of the embedding loss.

\section{Experiments}

Next we describe experimental evidence which demonstrates the performance of our method by advancing the state-of-the-art on the KITTI Multi-Object and Tracking Dataset~\citep{voigtlaender19}.

\subsection{Dataset}
The KITTI Multi-Object Tracking and Segmentation (MOTS) dataset contains 8,008 frames with instance segmentation labels resulting in a total of 26,899 annotated cars (see \Cref{table:kitti-dataset}). It is composed of 21 scenes with a resolution of $375\times 1242$ with consistent instance ID labels across time, allowing the training of video instance segmentation models. The frames are annotated at 10 frames per second, which is suitable for self-supervised monocular depth prediction. 

\begin{table}[h]
\centering
\begin{tabular}{l|ccccc}
\thickhline
 & Scenes & Frames & Annotations & Avg. \# frames & Avg. \# annotations\\
 \hline
Train & 12 & 5,027 & 18,831 & 419 & 1,569\\ 
Validation & 9 & 2,981 & 8,068 & 331 & 896\\
\thickhline
\end{tabular}
\caption{Details of the KITTI Multi-Object Tracking and Segmentation (MOTS) dataset.}
\label{table:kitti-dataset}
\end{table}

The ApolloScape dataset \citep{huang18} also contains video instance segmentation labels for 49,287 frames, but the annotations are not consistent in time, rendering the training of a temporal model impossible. NuScenes \citep{nuscenes19} features 1,000 scenes of 20 seconds with annotations at 2Hz in a diverse range of environments (different weather, daytime, city) but only contains bounding box labels, failing to represent the fine-grained details of instance segmentation. Temporal instance segmentation is also available on short snippets of the DAVIS dataset \citep{ponttuset17}, but each snippet is recorded by a different camera and is too short to effectively learn a depth model. For this reason, we focus on the KITTI MOTS dataset -- it is the only dataset that contains consistent video instance segmentation in a sufficient quantity to train deep models.

\subsection{Hyper-Parameters}
We halve the input images to our encoder to use an input RGB resolution of $192\times 640$. The spatio-temporal representation $z_t \in \mathbb{R}^{128\times24\times80}$ and the embedding dimension is $p=8$. Except for the experiments in \Cref{table:seq-len}, we train with a sequence length of 5 which corresponds to 0.5 seconds of temporal context since the videos are 10Hz.

In the loss function, we set the attraction radius $\rho_a=0.5$ and repulsion radius $\rho_r = 1.5$. We weight the losses with attraction and repulsion loss weight $\lambda_a = \lambda_r =1.0$, regularisation loss $\lambda_{\text{reg}} =0.001$ and depth loss $\lambda_{vs}=1.0$.

\subsection{Metrics}
In this section, we define multi-object tracking and segmentation metrics, measuring the quality of the segmentation as well as the consistency of the predictions over time. Let us denote by $\mathbb{H}$ the set of predicted ids, $\mathbb{G}$ the set of ground truth ids and $\psi$ the mapping from hypothesis segmentations to ground truth segmentations. $\psi: \mathbb{H} \to \mathbb{G} \cup \emptyset$ is defined as:

\begin{equation}
\psi(h) =
	\begin{cases}
	 \text{argmax}_g~\text{IoU}(h,g), & \text{if} \max_g \text{IoU}(h,g) > \text{threshold}\\
	 \emptyset, & \text{otherwise}
	\end{cases}
\end{equation}


We further define the following sets: $TP$ (true positives), $FP$ (false positives), $FN$ (false negatives), $IDS$ (the set of ID switches), and $\tilde{TP}$ the soft number of true positives: $\tilde{TP} = \sum_{h \in TP} \text{IoU}(h, \psi(h))$.

Following \citet{voigtlaender19}, we define the following MOTS metrics: multi-object tracking and segmentation precision (MOTSP), multi-object tracking and segmentation accuracy (MOTSA) and finally the soft multi-object tracking and segmentation accuracy (sMOTSA) that measures segmentation as well as detection and tracking quality.

\begin{align}
\text{MOTSP} &= \frac{|\tilde{TP}|}{|TP|} \\
\text{MOTSA} &= 1 - \frac{|FP| + |FN| + |IDS|}{|\mathbb{M}|} = \frac{|TP| - |FP|  - |IDS|}{|\mathbb{M}|} \\
\text{sMOTSA} &= \frac{|\tilde{TP}| - |FP| - |IDS|}{|\mathbb{M}|}
\end{align}

\subsection{Results}

We compare our model to the following baselines for video instance segmentation and report the results in \Cref{table:kitti-validation}.
\begin{itemize}[noitemsep,topsep=0pt]
	\item \textbf{Single-frame embedding loss}~\citep{brabandere17}, previous state-of-the-art method where instance segmentations are propagated in time using intersection-over-union association.
	\item \textbf{Mask R-CNN} \citep{he17}, instances are propagated with intersection-over-union.
	\item \textbf{Without temporal model}, spatio-temporal embedding loss, without the temporal model.
	\item \textbf{Without depth}, temporal model and spatio-temporal embedding loss, without the depth loss.
\end{itemize}

\begin{table*}[h]
\begin{adjustwidth}{-.5in}{-.5in}  
\centering
\begin{tabular}{lcccccccc}
\thickhline
& MOTSA & sMOTSA & MOTSP & AP \\ 
\hline
\citet{brabandere17} & 0.575 & 0.423  & 0.803 & 0.612 & \\ 
Mask R-CNN & 0.584 & 0.455  & \textbf{0.833} & \textbf{0.646} \\
\hline
Without temporal model & 0.582 & 0.426  & 0.799 & 0.607 \\ 
Without depth & 0.591 & 0.433  & 0.801 & 0.614 \\ 
\textbf{Ours} & \textbf{0.613} & \textbf{0.461} & 0.801 & 0.600 \\
\thickhline
\end{tabular}
\caption{KITTI MOTS validation set results comparing our model with baseline approaches.}
\label{table:kitti-validation}
\end{adjustwidth}
\end{table*}

The static detection metrics (MOTSP and average precision) are evaluated image by image without taking into account the temporal consistency of the instance segmentations. As the compared models (Without temporal model, Without depth, Ours) are all using the same mask network, they show similar performance in terms of detection.
However, when evaluating performance on metrics that measures temporal consistency (MOTSA and sMOTSA), our best model shows significant improvement over the baselines.

The variant without the temporal model performs poorly as it does not have any temporal context to learn a spatio-temporal embedding and therefore only relies on spatial appearance. The temporal model on the other hand learns with the temporal context and local motion, which results in a better embedding.
Our model, which learns to predict both a spatio-temporal embedding and monocular depth, achieves the best performance. In addition to using cues from appearance and temporal context, estimating depth allows the network to use information from the relative distance of objects to disambiguate them. Finally, we observe that our model outperforms Mask R-CNN \citep{he17} on the temporal metrics (MOTSA and sMOTSA) even though the latter exhibits a higher detection accuracy, further demonstrating the temporal consistency quality of our spatio-temporal embedding.

\subsubsection{Analysis of clustering and background segmentation.}
Our model relies on segmenting the background to determine the pixel locations to consider for instance clustering when applying mean shift. We evaluate the impact of using the ground truth mask against our predicted mask in \Cref{table:cluster-mask}. The performance gain is significant, hinting that our model could be improved with a more powerful mask network. 

Next, we evaluate the effect of clustering. In the best scenario, the validation loss would be zero, and the clustering would be perfect using the mean shift algorithm. However, this scenario is unlikely and the clustering algorithm is affected by noisy embeddings. We evaluate the effect of this noise by clustering with the ground-truth mean: we threshold with a radius $\rho_r$ around the ground truth instance embedding mean. This also results in a boost in the evaluation metrics, but most interestingly, a model that uses both ground truth instance embedding mean clustering and ground truth mask performs worse than a model using the ground truth mask and our clustering algorithm. This is because our clustering algorithm accumulates embeddings from past frames and therefore creates an attraction force for the mean shift algorithm that enables the instances to be matched more consistently.

\begin{table}[h]
\centering
\begin{tabular}{cccccccc}
\thickhline
GT Mean & GT Mask & MOTSA & sMOTSA & MOTSP & AP \\ 
\hline
\ding{55} & \ding{55}  & 0.613 & 0.461 & 0.801 & 0.600 \\ 
\ding{51} & \ding{55}  & 0.700 & 0.616 & 0.897 & 0.691 \\ 
\ding{55} & \ding{51}  & \textbf{0.804} & \textbf{0.751} & \textbf{0.936} & \textbf{0.786} \\ 
\ding{51} & \ding{51}  & 0.714 & 0.644 & 0.915 & 0.710 \\ 
\thickhline
\end{tabular}
\caption{Comparing the effect of noisy against ground-truth clustering and background segmentation on the KITTI MOTS dataset.}
\label{table:cluster-mask}
\end{table}

\subsubsection{Effect of the Sequence Length}

Our model learns a spatio-temporal embedding that enables clustering the video-pixels of each instance. Instance correspondence between frames is achieved by matching newly detected instances to previous instances if the mean embedding distance is below the repulsion radius, $p_r$. Therefore, We can track instances for an arbitrarily long period of time, as long as the embedding of a given instance changes smoothly over time, which is likely the case as temporal context and depth evolve progressively. 

However, when the network is trained over sequence of images which are too long, the learning process of the embedding collapses. This is because the attractive loss term is detrimental between distant frames as it pressures pixels from the same instance to have corresponding embeddings when their appearance and depth is no longer similar. It also suggests our model is able to reason over lower order motion cues more effectively than longer term dynamics. This is seen experimentally in \Cref{table:seq-len}.

\begin{table}[h]
\centering
\vspace{7mm}
\begin{tabular}{lcccccc}
\thickhline
Length & 1 & 3 & 5 & 7 & 10 & 15\\ 
\hline
MOTSA & 0.575 & 0.590 & \textbf{0.613} & 0.555 & 0.538 & 0.402\\
sMOTSA & 0.423 & 0.435 & \textbf{0.461} & 0.402 & 0.398 & 0.273\\
MOTSP & 0.803 & \textbf{0.810} & 0.801 & 0.788 & 0.792 & 0.783 \\
\thickhline
\end{tabular}
\caption{Influence of the sequence length on model performance. This indicates that our model can learn short-term motion features effectively, but not long-term cues. We reason that this is because over longer sequences, the loss prevents the embedding smoothly shifting, which naturally occurs to changing pose, appearance, context and lighting in the scene. We find the optimum sequence length on this dataset to be five.}
\label{table:seq-len}
\end{table}

\subsection{Qualitative examples}
\begin{figure}[h]
    \centering
    \begin{subfigure}[b]{\textwidth}
        \includegraphics[width=\linewidth]{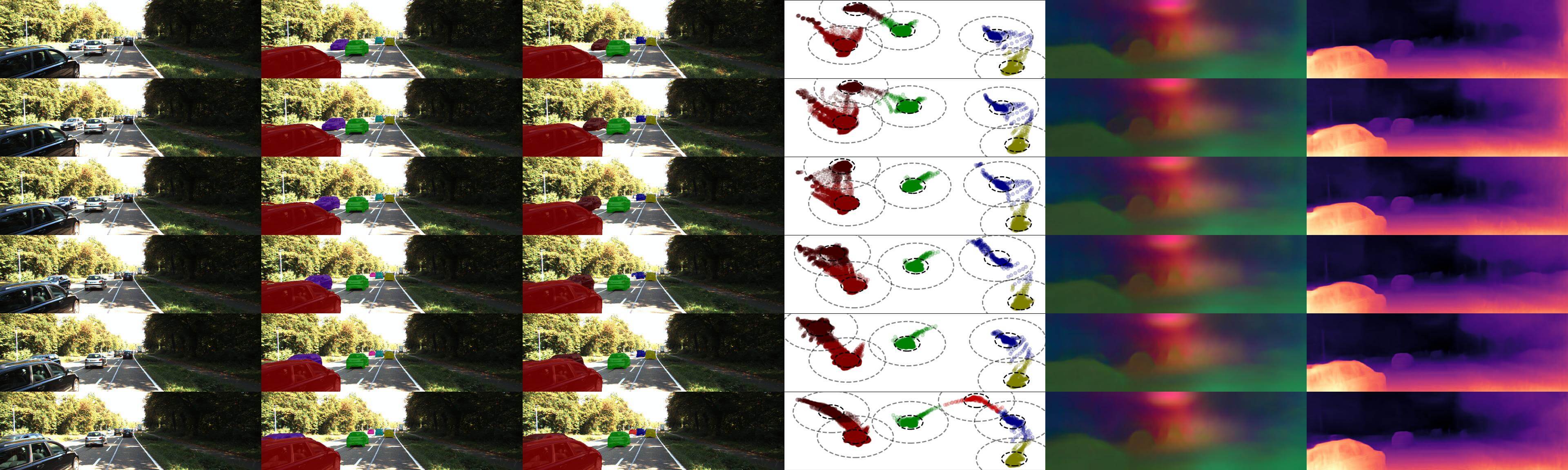}
        \caption{\textbf{Partial occlusion.} The brown car is correctly segmented even when being partially occluded by the red car, as the embedding contains past temporal context and is aware of the motion of the brown car.}
        \label{fig:partial-occlusion}
    \end{subfigure}
    \vskip\baselineskip
    \begin{subfigure}[b]{\textwidth}
        \includegraphics[width=\linewidth]{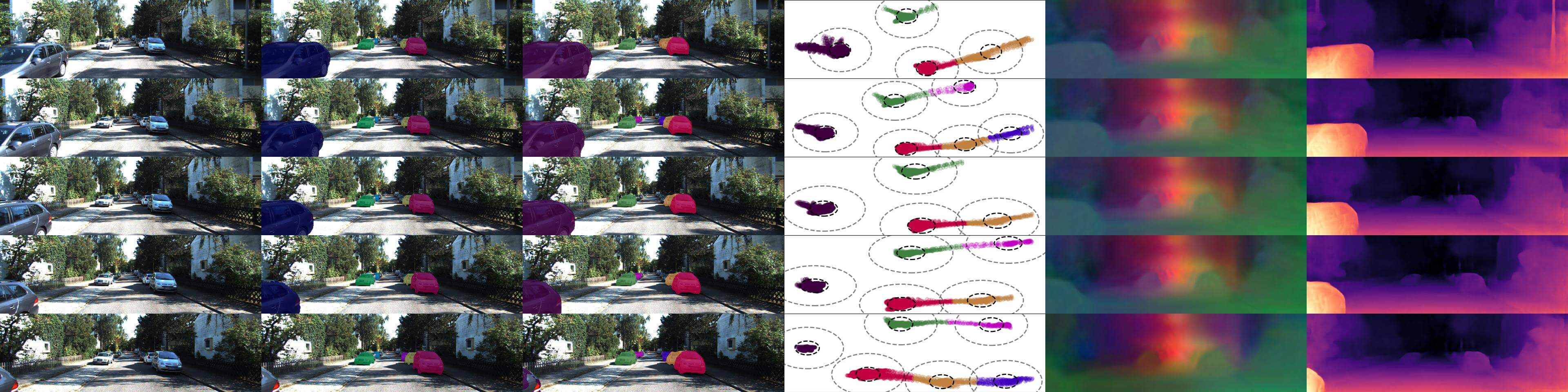}
        \caption{\textbf{Continuous tracking.} The segmented pink and purple cars are accurately tracked even with missing detections.}
        \label{fig:continuous-tracking}
    \end{subfigure}
    \vskip\baselineskip
    \begin{subfigure}[b]{\textwidth}
        \includegraphics[width=\linewidth]{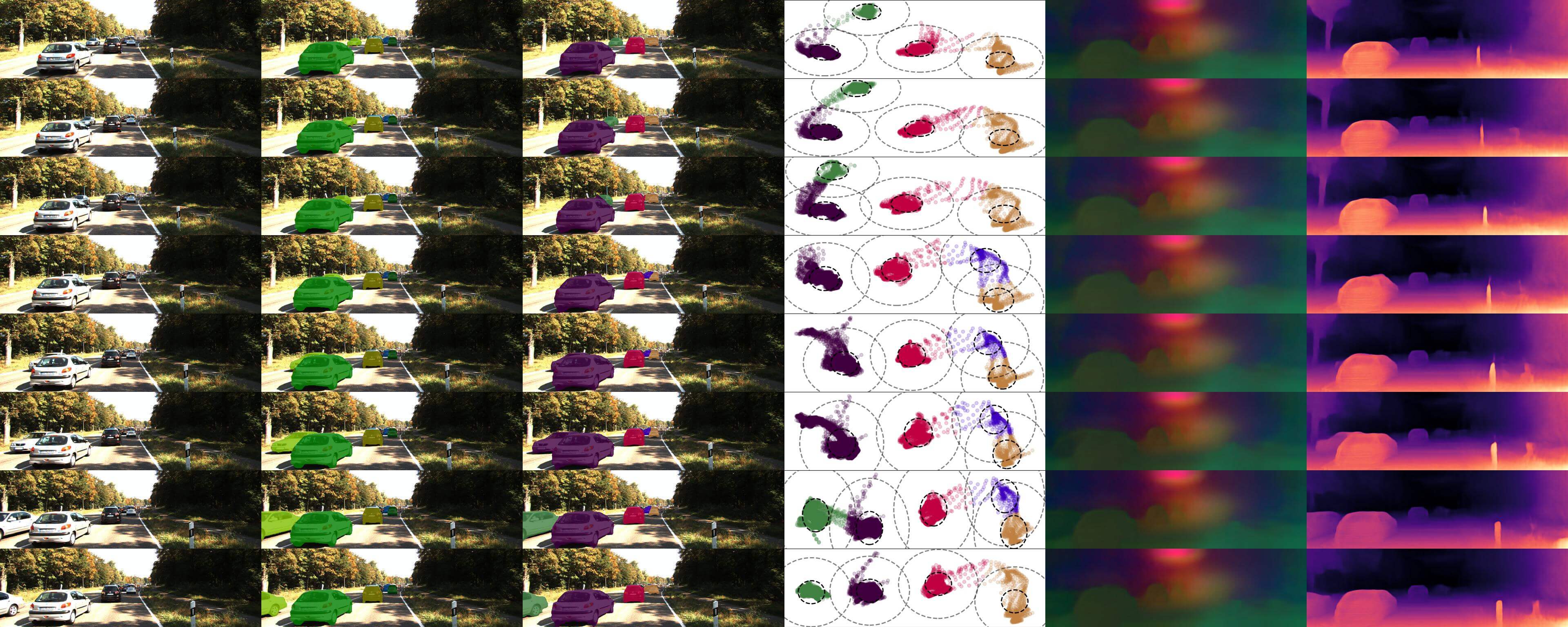}
        \caption{\textbf{Total occlusion.} The green car is correctly tracked, even though it was completely occluded by another car.}
        \label{fig:total-occlusion}
    \end{subfigure}
    \caption{From left to right: RGB input image, ground truth instance segmentation, predicted instance segmentation, embedding visualised in 2D, embedding visualised in RGB and predicted monocular depth.}
\end{figure}

We show that our model can consistently segment instances over time on the following challenging scenarios: tracking through partial (\Cref{fig:partial-occlusion} and full occlusion (\Cref{fig:total-occlusion}), and continuous tracking through noisy detections (\Cref{fig:continuous-tracking}).
Additional examples and failure cases of our model are shown in \Cref{appendix-qual} and in our \href{https://youtu.be/dc-3meFF6z0}{video demo}.

In each example, we show from left to right: RGB input image, ground truth instance segmentation, predicted instance segmentation, embedding visualised in 2D, embedding visualised in RGB and predicted monocular depth. The embedding is visualised in 2D and coloured with the results of the mean shift clustering. Each colour represents a different instance, the inner circle indicates the attraction radius from the instance mean embedding, and the outer circle represents the repulsion radius of each instance. Additionally, we also visualise the embedding spatially in 3D, by projecting its three principal components to an RGB image.

Finally, we show in \Cref{appendix-qual} that incorporating depth context greatly improves the quality of the embedding, especially in complex scenarios such as partial or total occlusion. We also observe that the embedding is much more structured when using depth, further validating that 3D geometry is essential to reason about dynamics agents in video.

\section{Conclusions}
We presented a new spatio-temporal embedding loss that generates consistent instance segmentation over time. The temporal network models the past temporal context and the depth network constrains the embedding to aid disambiguation between objects. We demonstrated that our model could effectively track occluded instances or instances with missed detections, by leveraging the temporal and depth context. Our method advanced the state-of-the-art at video instance segmentation on the KITTI Multi-Object and Tracking Dataset.

\clearpage

\bibliography{iclr2020_conference}
\bibliographystyle{iclr2020_conference}

\newpage
\appendix
\section{Appendix}
\label{appendix}

\subsection{Network Architecture Details}
\label{appendix-arch}
We report the details of each component of our model in this section. The number of parameters and layers of each module are in Table \ref{table:parameters}.
\begin{table}[h]
\centering
\begin{tabular}{lccccc}
\thickhline
 & Encoder & Temporal & Decoders & Pose & Mask\\ 
\hline
Parameters & 14.6M & 0.7M & 0.4M & 13.0M & 14.8M\\
Layers & 18 & 36 & 7 & 22 & 25\\
\thickhline
\end{tabular}
\caption{Number of parameters and layers of each module.}
\label{table:parameters}
\end{table}

\paragraph{Encoder.}
The encoder is a ResNet-18 convolutional layer \citep{he16}, with 128 output channels.

\paragraph{Temporal model.}
The temporal model contains 12 residual 3D convolutional blocks. Each residual block is the succession of: a 3D projection convolution with kernel size $1\times1\times1$ to halve the number of channels, a 3D causal convolutional layer with kernel $t\times3\times3$, and a 3D projection convolution with kernel $1\times1\times1$ to restore the number of channels. 

We set the temporal kernel size to $t=2$, and the number of output channels to 128.
\paragraph{Decoders.}
The decoders for instance embedding and depth estimation are identical and consist of 7 convolutional layers with channels [64, 64, 32, 32, 16, 16] and 3 upsampling layers. The final convolutional layer contains $p=8$ channels for instance embedding and $1$ channel for depth.

\paragraph{Depth Masking.}
During training, we remove from the photometric reprojection loss the pixels that violate the rigid scene assumption, i.e. the pixels whose appearance do not change between adjacents frames. We set the mask $M$ to only include pixels where the reprojection error is lower with the warped image $\hat{I}_{s\rightarrow t}$ than the unwarped source image $I_s$:
$$M = \left[\min_s e(I_t, \hat{I}_{s\rightarrow t}) < \min_s e(I_t, I_s)\right]$$

\paragraph{Pose Network.}
The pose network is the succession of a ResNet-18 model followed by 4 convolutions with [256, 256, 256, 6] channels. The last feature map is averaged to output a single 6-DoF transformation matrix.

\paragraph{Mask Network.}
The mask network is trained separately to mask the background and is the succession of the Encoder and Decoder described above.

\subsection{Additional Qualitative Examples}
\label{appendix-qual}

The following examples show qualitative results and failure examples of our video instance segmentation model on the KITTI Multi-Object and Tracking Dataset. From left to right: RGB input image, ground truth instance segmentation, predicted instance segmentation, embedding visualised in 2D, embedding visualised in RGB and predicted monocular depth.

\begin{figure}[h]
    \centering
    \begin{subfigure}[b]{\textwidth}
        \includegraphics[width=\linewidth]{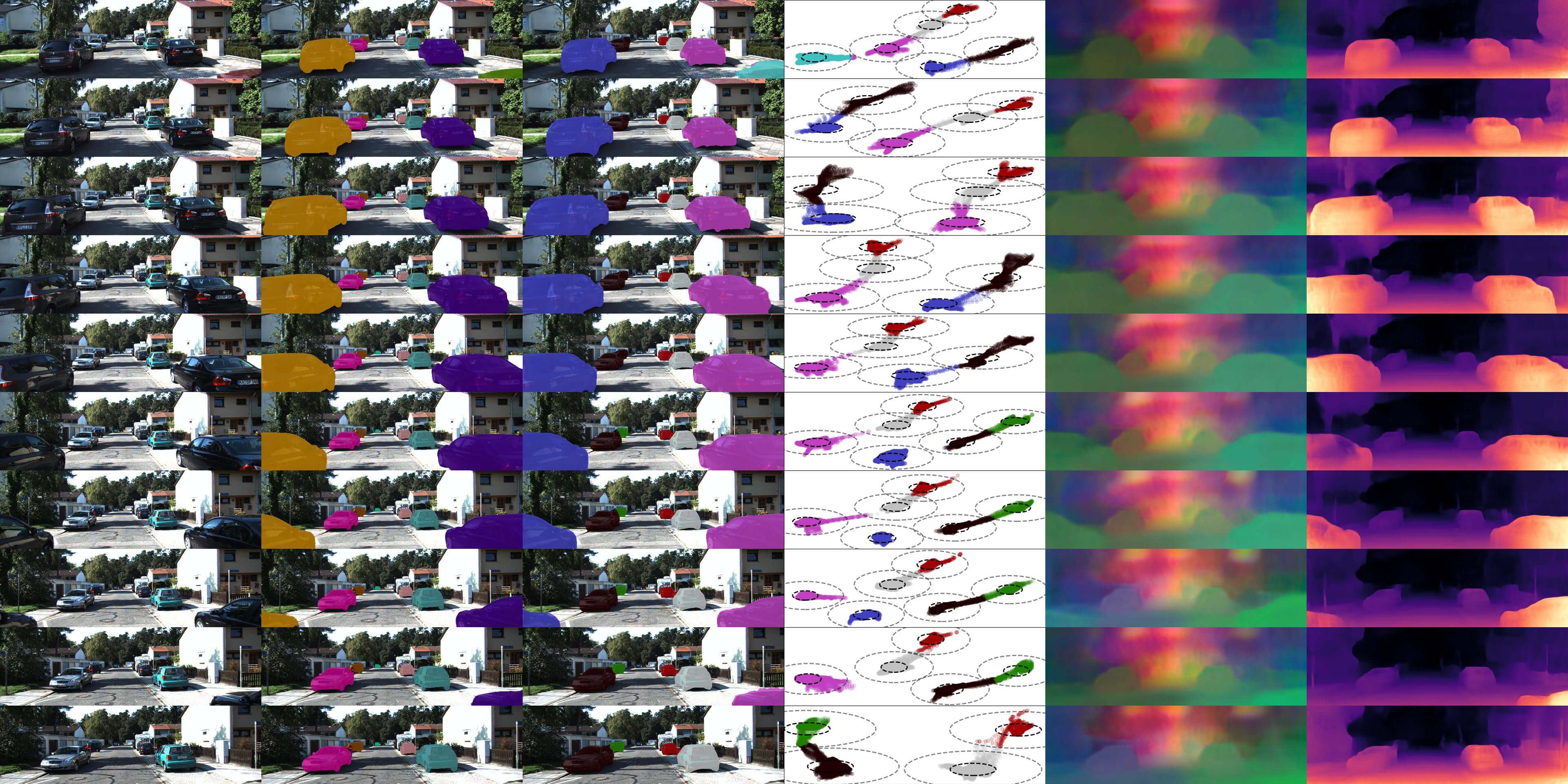}
        \caption{Video instance segmentation of parked cars.}
    \end{subfigure}
    \vskip\baselineskip
    \begin{subfigure}[b]{\textwidth}
        \includegraphics[width=\linewidth]{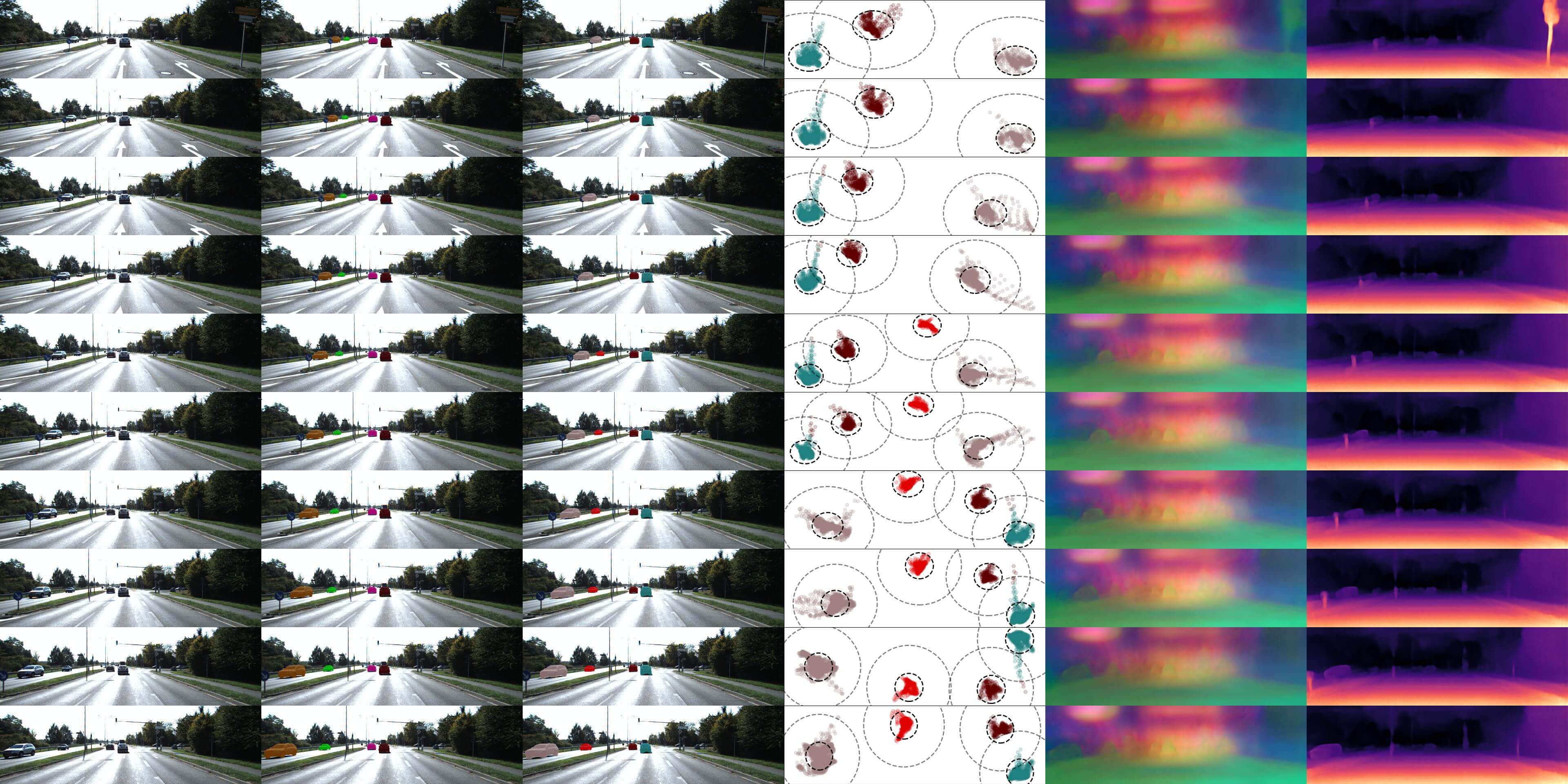}
        \caption{Video instance segmentation of other traffic.}
    \end{subfigure}
    \caption{From left to right: RGB input image, ground truth instance segmentation, predicted instance segmentation, embedding visualised in 2D, embedding visualised in RGB and predicted monocular depth.}
\end{figure}

\begin{figure}[h]
    \centering
    \begin{subfigure}[b]{\textwidth}
        \includegraphics[width=\linewidth]{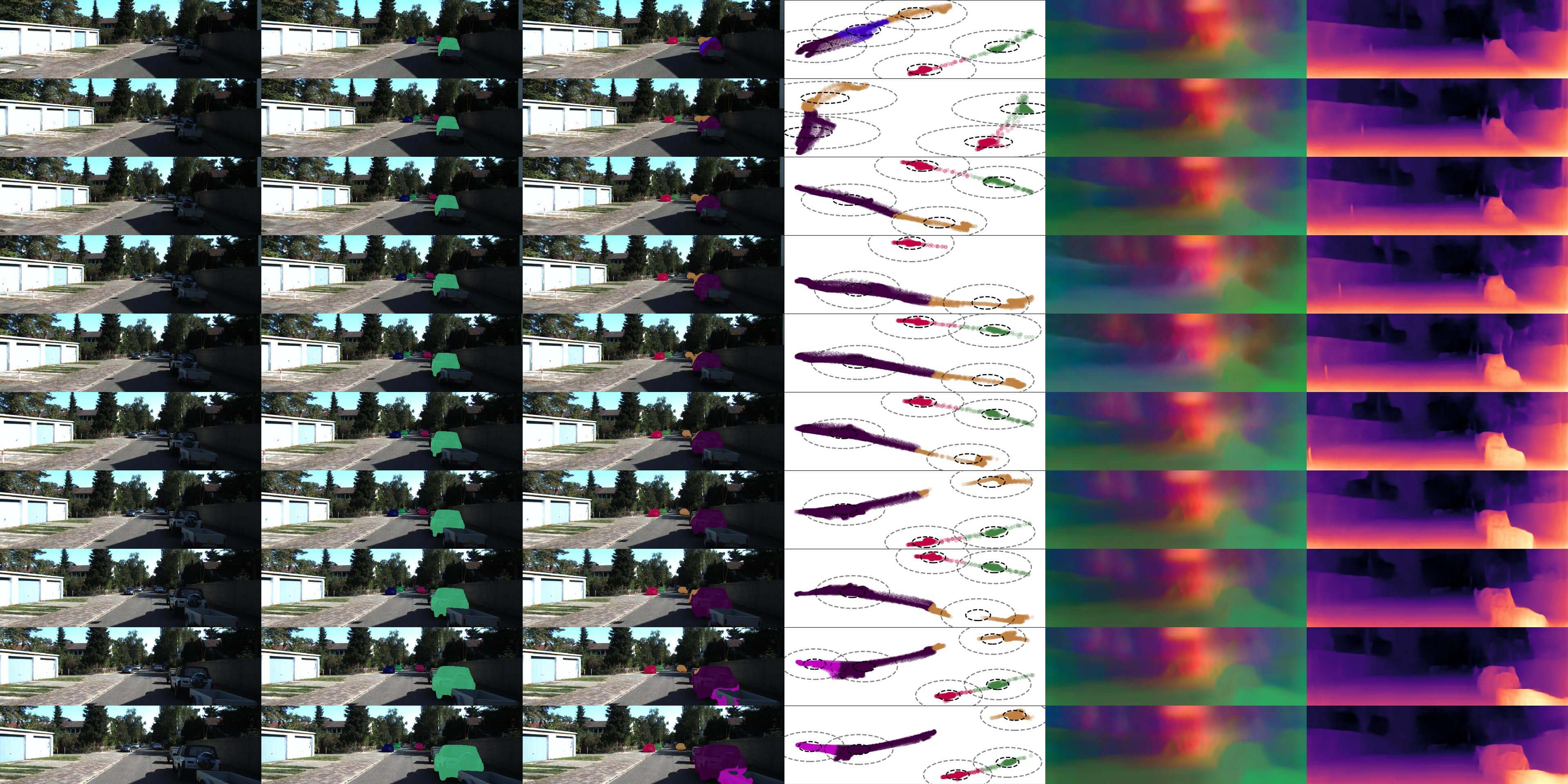}
        \caption{Failure case: the vehicle is segmented into two separate instances.}
    \end{subfigure}
    \vskip\baselineskip
    \begin{subfigure}[b]{\textwidth}
        \includegraphics[width=\linewidth]{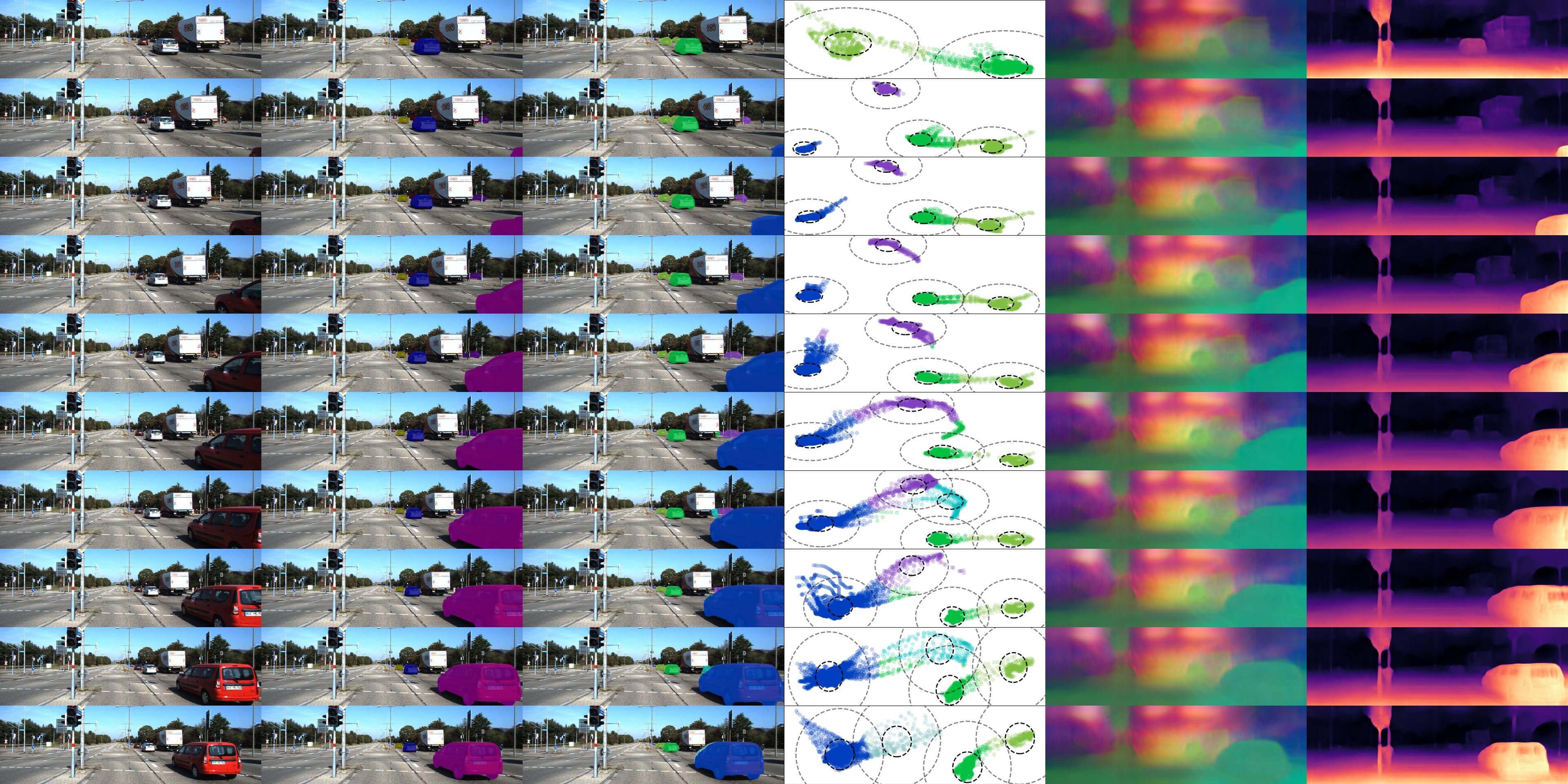}
        \caption{Failure case: two far-away cars are segmented as one instance.}
    \end{subfigure}
    \caption{From left to right: RGB input image, ground truth instance segmentation, predicted instance segmentation, embedding visualised in 2D, embedding visualised in RGB and predicted monocular depth.}
\end{figure}

\clearpage

\subsection{Qualitative Improvements with Depth}
\label{appendix-depth}
We show that our model greatly benefits from depth estimation, with the learned embedding being more structured, and correctly tracking objects in difficult scenarios such as partial or total occlusion.

\begin{figure}[h]
    \centering
    \begin{subfigure}[b]{\textwidth}
        \centering
        \includegraphics[width=\linewidth]{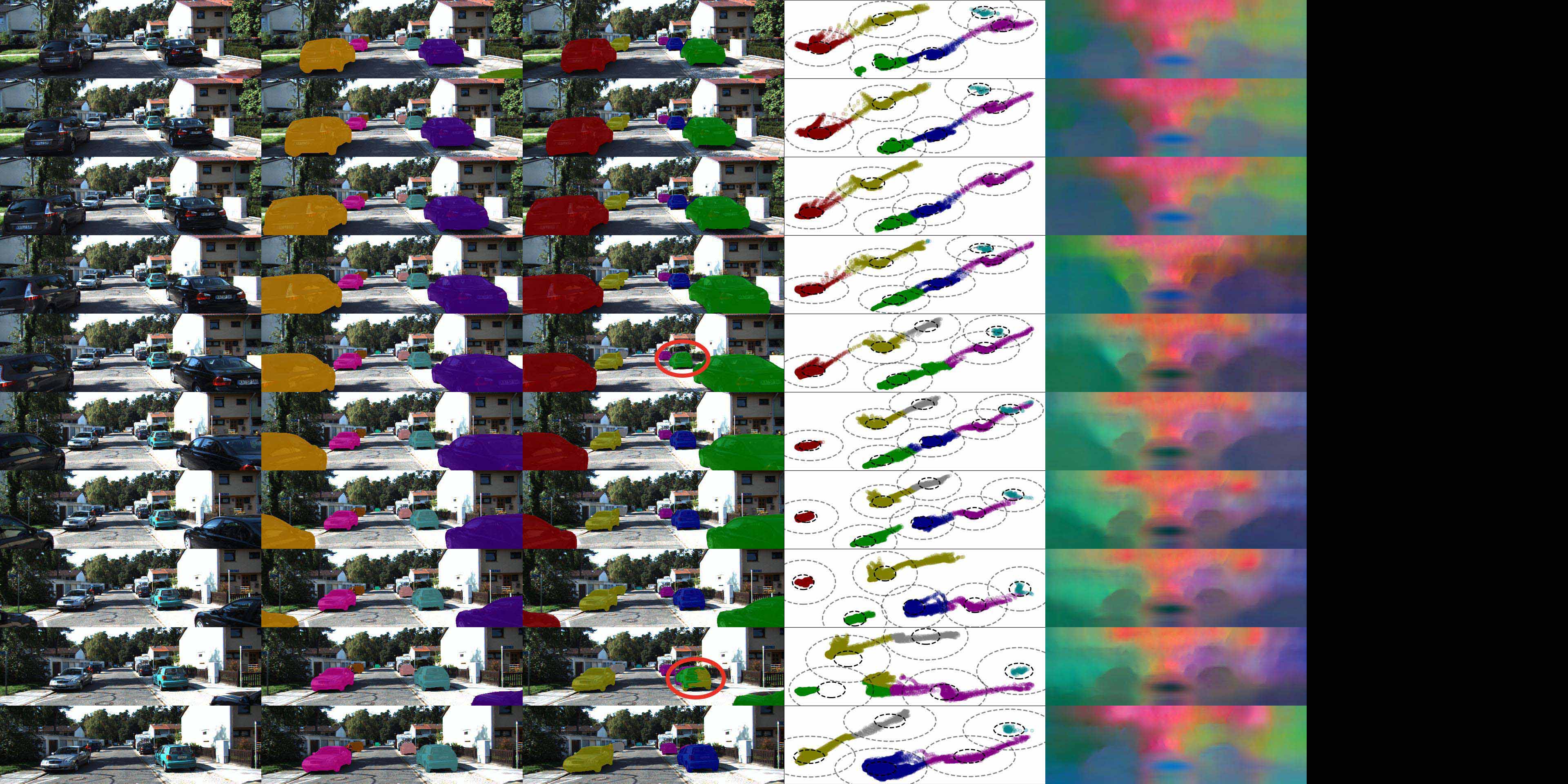}%
        \caption{Without depth estimation.}
    \end{subfigure}
    \vskip\baselineskip
    \begin{subfigure}[b]{\textwidth}
        \centering
        \includegraphics[width=\linewidth]{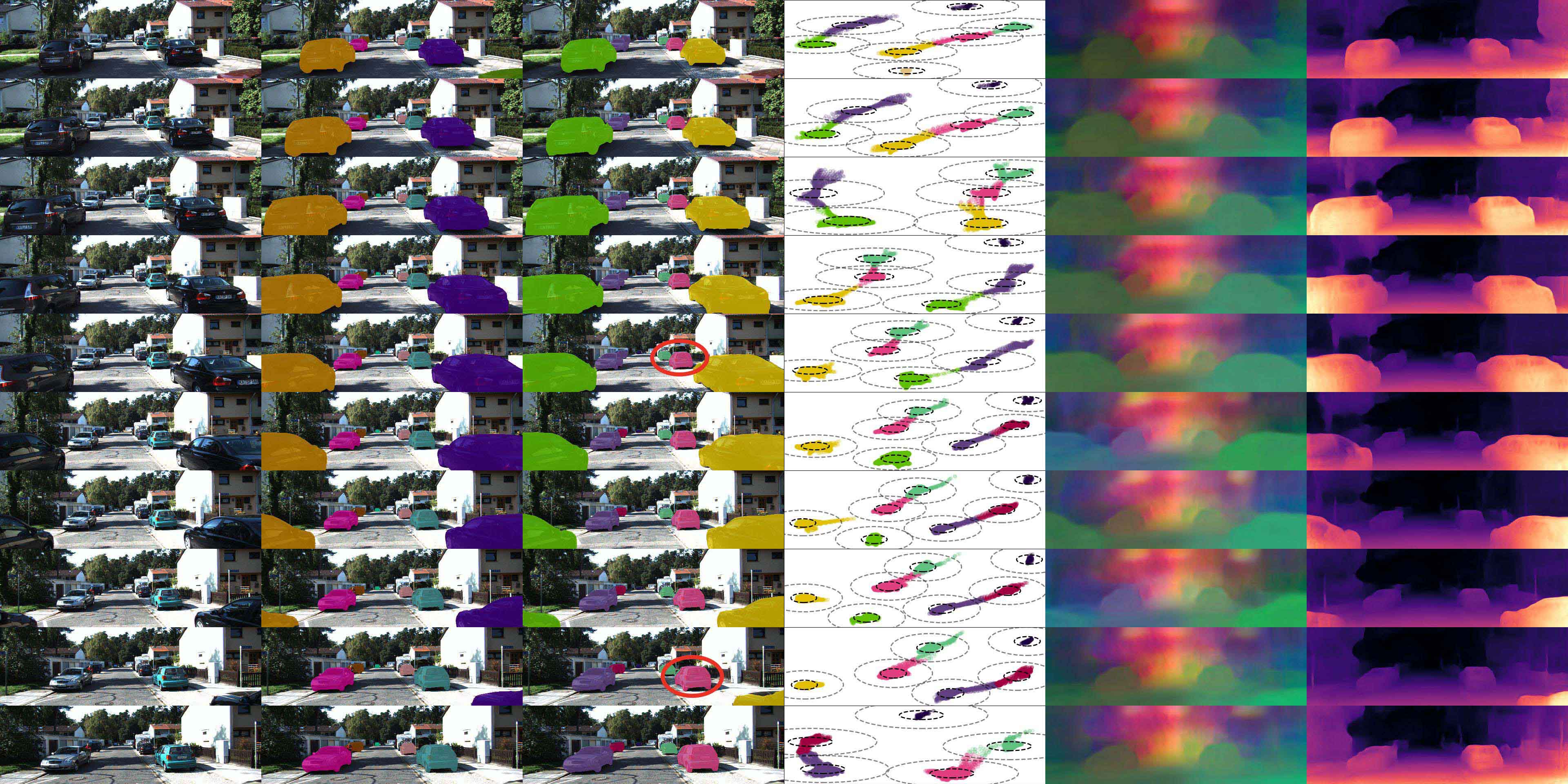}%
        \caption{With depth estimation}
    \end{subfigure}
    \caption{From left to right: RGB input image, ground truth instance segmentation, predicted instance segmentation, embedding visualised in 2D, embedding visualised in RGB and predicted monocular depth. Without depth estimation, the car circled in red is wrongly tracked in frame 5 and 9, while the model with depth correctly tracks it as the network has learned a consistent embedding based not only on appearance, but also on 3D geometry. Also, the RGB projection of the embedding from our model is considerably better and much more structured.}
\end{figure}

\begin{figure}[h]
    \centering
    \begin{subfigure}[b]{\textwidth}
        \centering
        \includegraphics[width=\linewidth]{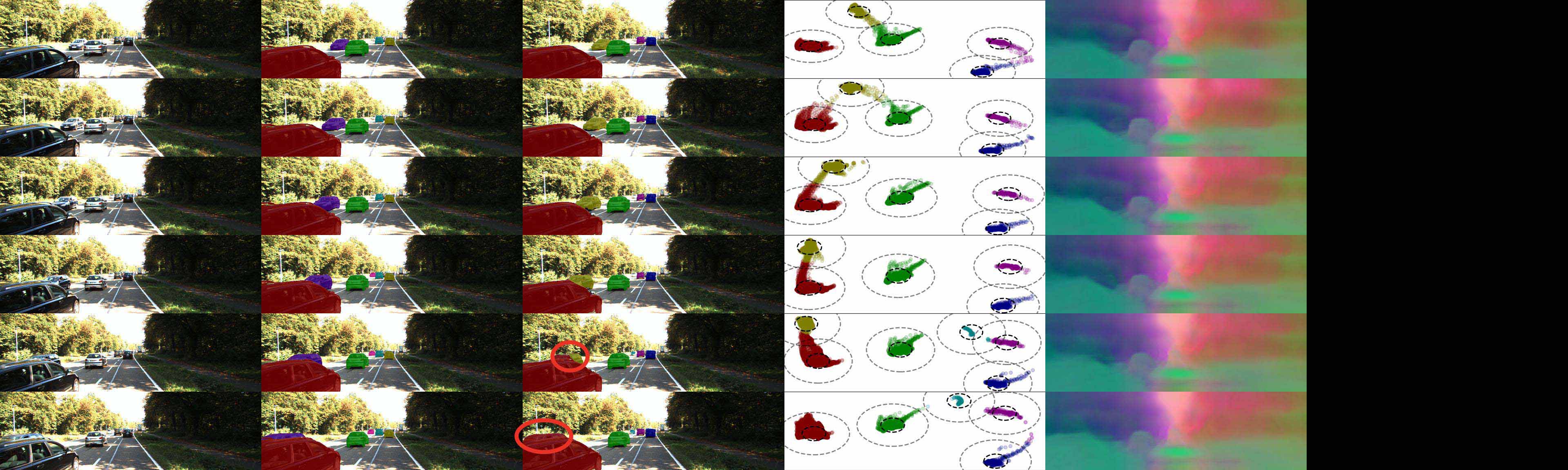}%
        \caption{Without depth estimation.}
    \end{subfigure}
    \vskip\baselineskip
    \begin{subfigure}[b]{\textwidth}
        \centering
        \includegraphics[width=\linewidth]{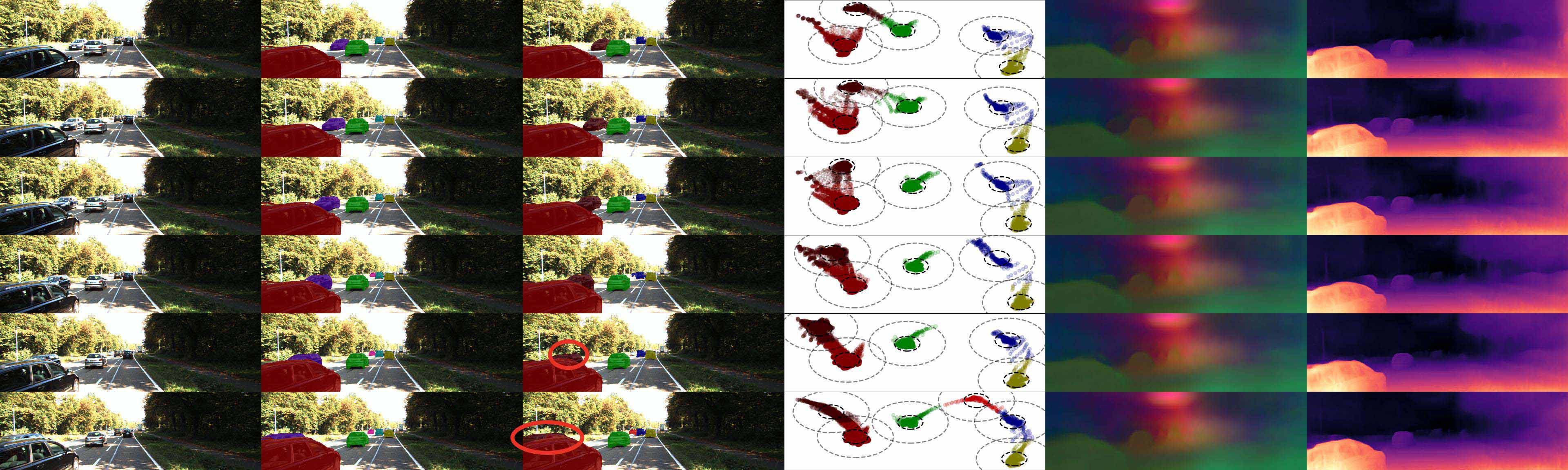}%
        \caption{With depth estimation}
    \end{subfigure}
    \caption{From left to right: RGB input image, ground truth instance segmentation, predicted instance segmentation, embedding visualised in 2D, embedding visualised in RGB and predicted monocular depth. Without depth estimation, the circled car merges into the red car, while the model with depth does not as there is a significant difference in depth between the two cars.}
\end{figure}

\begin{figure}[h]
    \centering
    \begin{subfigure}[b]{\textwidth}
        \centering
        \includegraphics[width=\linewidth]{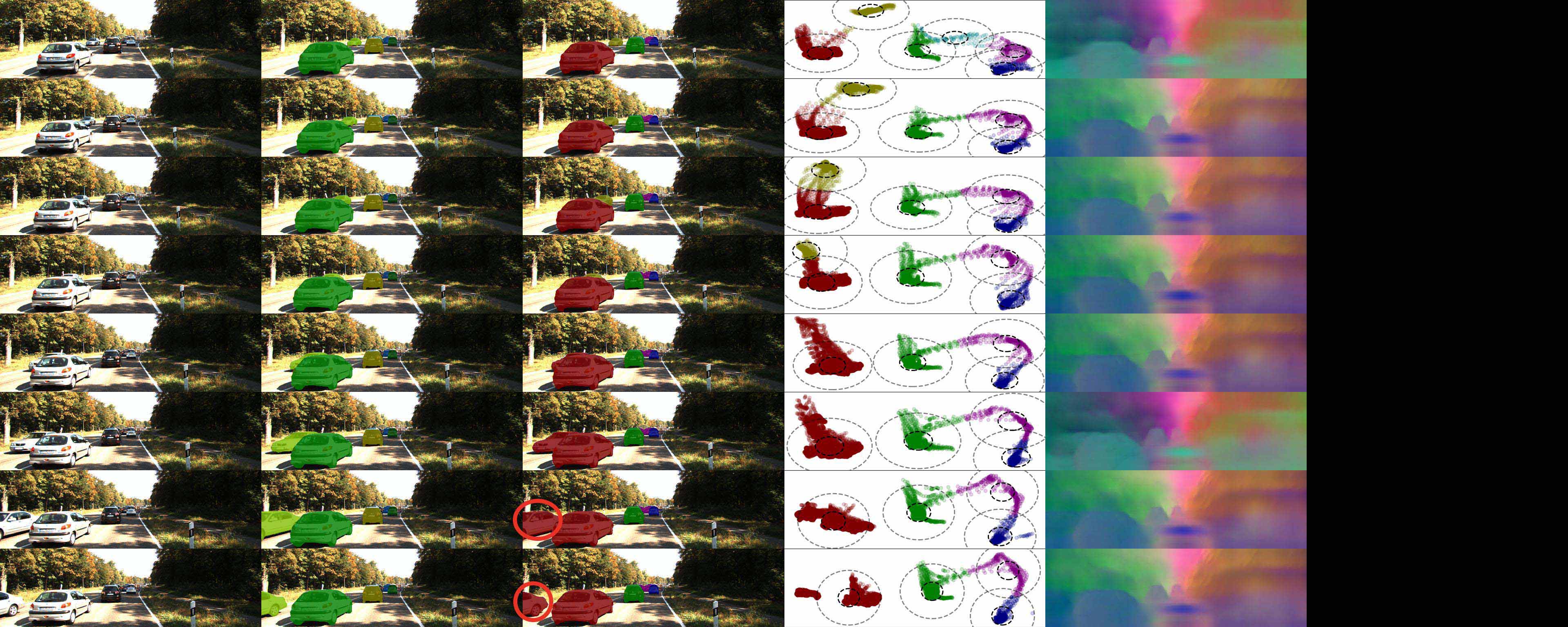}%
        \caption{Without depth estimation.}
    \end{subfigure}
    \vskip\baselineskip
    \begin{subfigure}[b]{\textwidth}
        \centering
        \includegraphics[width=\linewidth]{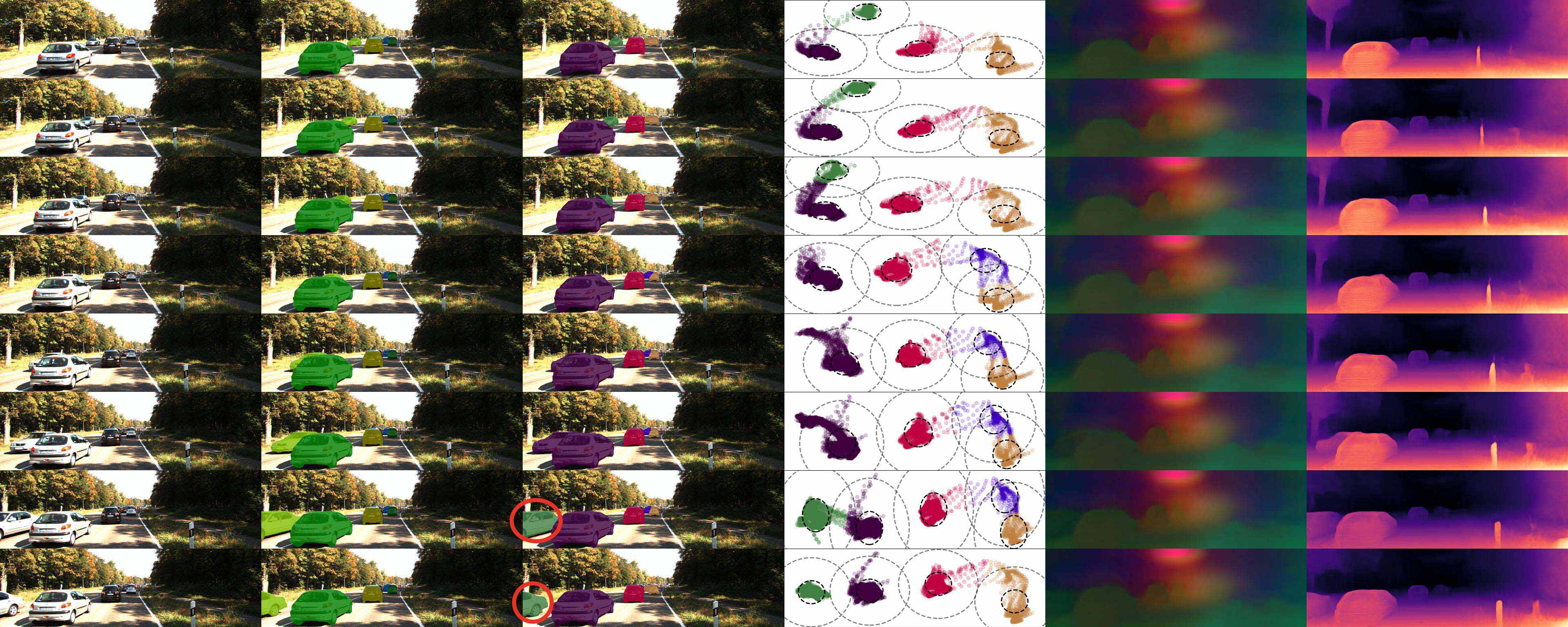}%
        \caption{With depth estimation}
    \end{subfigure}
    \caption{From left to right: RGB input image, ground truth instance segmentation, predicted instance segmentation, embedding visualised in 2D, embedding visualised in RGB and predicted monocular depth. The model without depth is not able to handle complete occlusion, while the model with depth can.}
\end{figure}

\end{document}

%% file: math_commands.tex
\usepackage{amsmath,amsfonts,bm}

\def\eqref#1{equation~\ref{#1}}

\def\1{\bm{1}}

\DeclareMathAlphabet{\mathsfit}{\encodingdefault}{\sfdefault}{m}{sl}
\SetMathAlphabet{\mathsfit}{bold}{\encodingdefault}{\sfdefault}{bx}{n}













%% file: iclr2020_conference.bbl
\begin{thebibliography}{30}
\providecommand{\natexlab}[1]{#1}
\providecommand{\url}[1]{\texttt{#1}}
\expandafter\ifx\csname urlstyle\endcsname\relax
  \providecommand{\doi}[1]{doi: #1}\else
  \providecommand{\doi}{doi: \begingroup \urlstyle{rm}\Url}\fi

\bibitem[Beyer et~al.(2017)Beyer, Breuers, Kurin, and Leibe]{beyer17}
Lucas Beyer, Stefan Breuers, Vitaly Kurin, and Bastian Leibe.
\newblock Towards a principled integration of multi-camera re-identification
  and tracking through optimal bayes filters.
\newblock In \emph{arXiv:1705.04608}, 2017.

\bibitem[Brabandere et~al.(2017)Brabandere, Neven, and Gool]{brabandere17}
Bert~De Brabandere, Davy Neven, and Luc~Van Gool.
\newblock Semantic instance segmentation with a discriminative loss function.
\newblock \emph{CVPRw}, abs/1708.02551, 2017.

\bibitem[Caesar et~al.(2019)Caesar, Bankiti, Lang, Vora, Liong, Xu, Krishnan,
  Pan, Baldan, and Beijbom]{nuscenes19}
Holger Caesar, Varun Bankiti, Alex~H. Lang, Sourabh Vora, Venice~Erin Liong,
  Qiang Xu, Anush Krishnan, Yu~Pan, Giancarlo Baldan, and Oscar Beijbom.
\newblock nuscenes: A multimodal dataset for autonomous driving.
\newblock \emph{arXiv preprint arXiv:1903.11027}, 2019.

\bibitem[Chen et~al.(2018)Chen, Hermans, Papandreou, Schroff, Wang, and
  Adam]{chen17b}
Liang-Chieh Chen, Alexander Hermans, George Papandreou, Florian Schroff, Peng
  Wang, and Hartwig Adam.
\newblock Masklab: Instance segmentation by refining object detection with
  semantic and direction features.
\newblock \emph{2018 IEEE/CVF Conference on Computer Vision and Pattern
  Recognition}, 2018.

\bibitem[Comaniciu \& Meer(2002)Comaniciu and Meer]{comaniciu2002mean}
Dorin Comaniciu and Peter Meer.
\newblock Mean shift: A robust approach toward feature space analysis.
\newblock \emph{IEEE Transactions on Pattern Analysis \& Machine Intelligence},
  pp.\  603--619, 2002.

\bibitem[Cordts et~al.(2016)Cordts, Omran, Ramos, Rehfeld, Enzweiler, Benenson,
  Franke, Roth, and Schiele]{cityscapes16}
Marius Cordts, Mohamed Omran, Sebastian Ramos, Timo Rehfeld, Markus Enzweiler,
  Rodrigo Benenson, Uwe Franke, Stefan Roth, and Bernt Schiele.
\newblock The cityscapes dataset for semantic urban scene understanding.
\newblock In \emph{Proc. of the IEEE Conference on Computer Vision and Pattern
  Recognition (CVPR)}, 2016.

\bibitem[Fathi et~al.(2017)Fathi, Wojna, Rathod, Wang, Song, Guadarrama, and
  Murphy]{fathi17}
Alireza Fathi, Zbigniew Wojna, Vivek Rathod, Peng Wang, Hyun~Oh Song, Sergio
  Guadarrama, and Kevin~P. Murphy.
\newblock Semantic instance segmentation via deep metric learning.
\newblock \emph{ArXiv}, abs/1703.10277, 2017.

\bibitem[Godard et~al.(2019)Godard, {Mac Aodha}, Firman, and Brostow]{godard19}
Cl{\'{e}}ment Godard, Oisin {Mac Aodha}, Michael Firman, and Gabriel~J.
  Brostow.
\newblock Digging into self-supervised monocular depth prediction.
\newblock \emph{The International Conference on Computer Vision (ICCV)},
  October 2019.

\bibitem[He et~al.(2016)He, Zhang, Ren, and Sun]{he16}
Kaiming He, Xiangyu Zhang, Shaoqing Ren, and Jian Sun.
\newblock Deep residual learning for image recognition.
\newblock In \emph{CVPR}, 2016.

\bibitem[He et~al.(2017)He, Gkioxari, Dollar, and Girshick]{he17}
Kaiming He, Georgia Gkioxari, Piotr Dollar, and Ross Girshick.
\newblock {Mask R-CNN}.
\newblock In \emph{Proceedings of the International Conference on Computer
  Vision ({ICCV})}, 2017.

\bibitem[Hu et~al.(2018)Hu, Huang, and Schwing]{hu17}
Yuan-Ting Hu, Jia-Bin Huang, and Alexander~G. Schwing.
\newblock Maskrnn: Instance level video object segmentation.
\newblock \emph{NIPS}, abs/1803.11187, 2018.

\bibitem[Hu et~al.(2019)Hu, Chen, Hui, Huang, and Schwing]{hu19}
Yuan-Ting Hu, Hong-Shuo Chen, Kexin Hui, Jia-Bin Huang, and Alexander~G.
  Schwing.
\newblock Sail-vos: Semantic amodal instance level video object segmentation -
  a synthetic dataset and baselines.
\newblock In \emph{CVPR}, 2019.

\bibitem[Huang et~al.(2018)Huang, Cheng, Geng, Cao, Zhou, Wang, Lin, and
  Yang]{huang18}
Xinyu Huang, Xinjing Cheng, Qichuan Geng, Binbin Cao, Dingfu Zhou, Peng Wang,
  Yuanqing Lin, and Ruigang Yang.
\newblock The apolloscape dataset for autonomous driving.
\newblock In \emph{The IEEE Conference on Computer Vision and Pattern
  Recognition (CVPR) Workshops}, June 2018.

\bibitem[Jaderberg et~al.(2015)Jaderberg, Simonyan, Zisserman, and
  Kavukcuoglu]{jaderberg15}
Max Jaderberg, Karen Simonyan, Andrew Zisserman, and Koray Kavukcuoglu.
\newblock Spatial transformer networks.
\newblock In \emph{Proceedings of the 28th International Conference on Neural
  Information Processing Systems - Volume 2}, NIPS'15, pp.\  2017--2025,
  Cambridge, MA, USA, 2015. MIT Press.

\bibitem[Kendall et~al.(2018)Kendall, Gal, and Cipolla]{kendall2017multi}
Alex Kendall, Yarin Gal, and Roberto Cipolla.
\newblock Multi-task learning using uncertainty to weigh losses for scene
  geometry and semantics.
\newblock In \emph{Proceedings of the IEEE Conference on Computer Vision and
  Pattern Recognition ({CVPR})}, 2018.

\bibitem[Kong \& Fowlkes(2018)Kong and Fowlkes]{kong18}
Shu Kong and Charless Fowlkes.
\newblock Recurrent pixel embedding for instance grouping.
\newblock In \emph{2018 Conference on Computer Vision and Pattern Recognition
  (CVPR)}, 2018.

\bibitem[Lin et~al.(2014)Lin, Maire, Belongie, Bourdev, Girshick, Hays, Perona,
  Ramanan, Doll{\'a}r, and Zitnick]{coco14}
Tsung-Yi Lin, Michael Maire, Serge~J. Belongie, Lubomir~D. Bourdev, Ross~B.
  Girshick, James Hays, Pietro Perona, Deva Ramanan, Piotr Doll{\'a}r, and
  C.~Lawrence Zitnick.
\newblock Microsoft coco: Common objects in context.
\newblock In \emph{ECCV}, 2014.

\bibitem[Liu et~al.(2018)Liu, Qi, Qin, Shi, and Jia]{liu18}
Shu Liu, Lu~Qi, Haifang Qin, Jianping Shi, and Jiaya Jia.
\newblock Path aggregation network for instance segmentation.
\newblock \emph{2018 IEEE/CVF Conference on Computer Vision and Pattern
  Recognition}, pp.\  8759--8768, 2018.

\bibitem[Neuhold et~al.(2017)Neuhold, Ollmann, Rota~Bul\`o, and
  Kontschieder]{mapillary17}
Gerhard Neuhold, Tobias Ollmann, Samuel Rota~Bul\`o, and Peter Kontschieder.
\newblock The mapillary vistas dataset for semantic understanding of street
  scenes.
\newblock In \emph{International Conference on Computer Vision (ICCV)}, 2017.

\bibitem[Oord et~al.(2016)Oord, Dieleman, Zen, Simonyan, Vinyals, Graves,
  Kalchbrenner, Senior, and Kavukcuoglu]{oord16}
Aaron van~den Oord, Sander Dieleman, Heiga Zen, Karen Simonyan, Oriol Vinyals,
  Alex Graves, Nal Kalchbrenner, Andrew Senior, and Koray Kavukcuoglu.
\newblock Wavenet: A generative model for raw audio, 2016.

\bibitem[Paine et~al.(2016)Paine, Khorrami, Chang, Zhang, Ramachandran,
  Hasegawa-Johnson, and Huang]{paine16}
Tom~Le Paine, Pooya Khorrami, Shiyu Chang, Yang Zhang, Prajit Ramachandran,
  Mark~A Hasegawa-Johnson, and Thomas~S Huang.
\newblock Fast wavenet generation algorithm.
\newblock \emph{arXiv preprint arXiv:1611.09482}, 2016.

\bibitem[Pont-Tuset et~al.(2017)Pont-Tuset, Perazzi, Caelles, Arbelaez,
  Sorkine-Hornung, and {Van Gool}]{ponttuset17}
J.~Pont-Tuset, F.~Perazzi, S.~Caelles, P.~Arbelaez, A.~Sorkine-Hornung, and
  L.~{Van Gool}.
\newblock The 2017 davis challenge on video object segmentation.
\newblock \emph{arXiv:1704.00675}, 2017.

\bibitem[Sadeghian et~al.(2017)Sadeghian, Alahi, and Savarese]{sadeghian17}
Amir Sadeghian, Alexandre Alahi, and Silvio Savarese.
\newblock Tracking the untrackable: Learning to track multiple cues with
  long-term dependencies.
\newblock In \emph{ICCV}, 2017.

\bibitem[Son et~al.(2017)Son, Baek, Cho, and Han]{son17}
Jeany Son, Mooyeol Baek, Minsu Cho, and Bohyung Han.
\newblock Multi-object tracking with quadruplet convolutional neural networks.
\newblock \emph{2017 IEEE Conference on Computer Vision and Pattern Recognition
  (CVPR)}, pp.\  3786--3795, 2017.

\bibitem[Voigtlaender et~al.(2019)Voigtlaender, Krause, Osep, Luiten, Sekar,
  Geiger, and Leibe]{voigtlaender19}
Paul Voigtlaender, Michael Krause, Aljosa Osep, Jonathon Luiten, Berin
  Balachandar~Gnana Sekar, Andreas Geiger, and Bastian Leibe.
\newblock Mots: Multi-object tracking and segmentation.
\newblock In \emph{Proceedings IEEE Conf. on Computer Vision and Pattern
  Recognition (CVPR)}, 2019.

\bibitem[Vondrick et~al.(2018)Vondrick, Shrivastava, Fathi, Guadarrama, and
  Murphy]{vondrick18}
Carl~Martin Vondrick, Abhinav Shrivastava, Alireza Fathi, Sergio Guadarrama,
  and Kevin Murphy.
\newblock Tracking emerges by colorizing videos.
\newblock In \emph{ECCV}, 2018.

\bibitem[Xiang et~al.(2015)Xiang, Alahi, and Savarese]{xiang15}
Yu~Xiang, Alexandre Alahi, and Silvio Savarese.
\newblock Learning to track: Online multi-object tracking by decision making.
\newblock In \emph{International Conference on Computer Vision}, 2015.

\bibitem[Yang et~al.(2019)Yang, Fan, and Xu]{yang19}
Linjie Yang, Yuchen Fan, and Ning Xu.
\newblock Video instance segmentation.
\newblock \emph{ICCV}, 2019.

\bibitem[Zhao et~al.(2017)Zhao, Gallo, Frosio, and Kautz]{zhao17}
Hang Zhao, Orazio Gallo, Iuri Frosio, and Jan Kautz.
\newblock Loss functions for image restoration with neural networks.
\newblock \emph{IEEE Transactions on Computational Imaging}, 3:\penalty0
  47--57, 2017.

\bibitem[Zhou et~al.(2017)Zhou, Brown, Snavely, and Lowe]{zhou17}
Tinghui Zhou, Matthew Brown, Noah Snavely, and David Lowe.
\newblock Unsupervised learning of depth and ego-motion from video.
\newblock In \emph{Computer Vision and Pattern Recognition}, 2017.

\end{thebibliography}
